\documentclass[9pt]{IEEEtran}
\usepackage{spconf,amsmath,epsfig}
\usepackage{booktabs,multirow}
\usepackage{tabulary}
\usepackage{algpseudocode,algorithm,algorithmicx}
\usepackage{subfigure}
\usepackage{graphicx}
\usepackage{array}
\usepackage{amssymb}
\usepackage{amsmath}
\newcolumntype{P}[1]{>{\centering\arraybackslash}p{#1}}
\newcolumntype{M}[1]{>{\centering\arraybackslash}m{#1}}
\usepackage{amsmath,epsfig}
\usepackage{cite}
\usepackage{booktabs,multirow}
\usepackage{epstopdf}
\usepackage{color}
\usepackage[numbers,sort&compress]{natbib}
\usepackage[ocgcolorlinks]{hyperref}
\hypersetup{
    colorlinks,
    citecolor=black,
    filecolor=black,
    linkcolor=red,
    urlcolor=blue
}

\usepackage{mathrsfs}

\newcolumntype{K}[1]{>{\centering\arraybackslash}p{#1}}

\renewcommand{\refname}{\centerline{6.~~REFERENCES}}
\makeatletter
\renewcommand\bibsection{%
  \section*{{\normalsize{\refname}}\@mkboth{\MakeUppercase{\refname}}{\MakeUppercase{\refname}}}\vspace*{-.30\baselineskip}%
}%
\makeatother

\title{Title}
\title{Hierarchical Variational Autoencoders for visual Counterfactuals
}
\name{Nicolas Vercheval$^{1,2}$ and Aleksandra Pi\v zurica$^{1}$}
\address{$^{1}$Department of Telecommunications and Information Processing, TELIN-GAIM, \\Faculty of Engineering and Architecture, Ghent University, Belgium\\
$^{2}$Department of Electronics and Information Systems, Clifford Research Group, \\Faculty of Engineering and Architecture, Ghent University, Belgium}

\begin{document}

\maketitle
\begin{abstract}
Conditional Variational Auto Encoders (VAE) are gathering significant attention as an Explainable Artificial Intelligence (XAI) tool.
The codes in the latent space provide a theoretically sound way to produce counterfactuals, i.e. alterations resulting from an intervention on a targeted semantic feature. To be applied on real images more complex models are needed, such as Hierarchical CVAE. This comes with a challenge as the naive conditioning is no longer effective.  In this paper we show how relaxing the effect of the posterior leads to successful counterfactuals and we introduce VAEX\footnote{\url{https://github.com/nverchev/VAEX}} an Hierarchical VAE designed for this approach  that can visually audit a classifier in applications.
\end{abstract}


\section{Introduction}
Explainable artificial intelligence (XAI), considered an emerging research field for years, has now fully established itself as a major discipline and is able to answer the urgent need of interpretability and confidence required by international laws, consumers and final users. The guidelines~\cite{EU} from the European Commission encourage algorithms to be designed as an accurate guiding tool which still leaves the human evaluation at the center of the decision making process; humans must be aware of the AI System's limitations and able to determine its reliability, fairness and bias. In that respect, they expressly attribute a key role to auditing, which depending on the data at hand, might be done visually.

Visual perceptive interpretability has been flourishing in recent years with major breakthroughs\cite{Sundararajan,Ribeiro,Lundberg} in identifying objects or image components that were decisive for a given outcome of the automated analysis. Commonly the user is assumed to immediately understand why the highlighted objects determine the evaluation of the System. This solution does not work when, for example, the shape or the color of the object, rather than its presence, is the main reason for a prediction.

Class-contrastive \emph{counterfactuals}~\cite{Pearl}, realistic variations of a sample resulting from an hypothetical intervention on some of its variables, establish a causal relationship between an image and its evaluation from an investigated classifier, displaying a ``modus tollens'' inference by mirroring the human imagination, and delineate a causal link between different scenarios and the outcome \cite{Byrne}. The prior knowledge of the likelihood of the factors helps instead identifying the right causal structure from the explanation~\cite{Lara}. A good counterfactual~\cite{Keane} must be plausible, and different to the original sample only in a few key factors meaningful to an expert such as a medical doctor~\cite {Tjoa} or in sensitive attributes to test the network's fairness~\cite{Johnson}.

Conditional Variational Auto Encoders (VAE)~\cite{Kingma, Sohn} offer a perfect framework to create counterfactuals, as they are able to capture and disentangle latent representations connected to the targeted variable while the known prior distribution of their latent space ensures the plausibility of their hypothetical sample. Their use cases in XAI range from the design of metamaterials~\cite{Ma}, text prediction~\cite{Alvarez-Melis}, tabular data~\cite{Pawelczyk}, treatment selection~\cite{Louizos}~ and fairness in clinical predictions~\cite{Pfohl} but are nevertheless mostly limited to quantitative data.
As of now, VAE have been used to counterfeit only low-resolution images~\cite{Ma,Sadiq}.  One reason for it is that the reconstruction quality is traded off with the expressivity of the generative model~\cite{Higgins}, particularly in images of higher quality.
We can observe this in~\cite{Besserve}, where the authors introduce a technique to hybridize two generated samples, even from different datasets. Their VAE allow them to quickly apply their technique to real faces, but they struggle with reconstructing such dataset. 
Hierarchical VAE reduce this trade-off and are able to produce sharp images~\cite{Vahdat} but we show that directly conditioning the codes of a sample is no longer effective: the posterior distributions of detailed images typically lie in areas of the latent space with low density, where the information of the shallow layers overtakes any change in the deeper ones. 

At the same time, moving along the manifold the generative part establishes a reliable pseudo-distance which is not constrained by pixel loss.
This is crucial to improve over recent works such as~\cite{Barredo}, where they use autoencoders with adversarial loss conditioned on accessory attributes~\cite{He} and do a search to minimize the required perturbation that results in a different evaluation from an investigated classifier. Their similarity metric between samples relies on the mean square error and limits the expressivity of the counterfactual, making the change less intuitive and easily subject to bias. 

Hierarchical models are key for VAE to have sharp reconstructions, and explanatory methods that naturally interact with the hierarchical structure need to be investigated. The contribution of this paper is the following:
we introduce VAEX, a hierarchical Conditional VAE model which stresses a deep encoding of the images by using the statistics of the previous latent variables during inference and cascading in this way, the initial condition through the whole latent space.
Differently from previous work, VAEX is directly conditioned on the evaluation of a classifier to specifically target the bias of the classifier and to allow evaluation during the test phase. Furthermore, we do not need to iterate the classifier's evaluation, which can be long and complicated when the latter is an ensemble model, but we only make use of the probabilities that it associates to the samples.
Finally, we introduce a simple, quick and very effective way to create realistic sharp counterfactuals with this setup by taking advantage of the expressivity of the generative model in an end to end way. This shows how the gradual change of a sample image reliably leads to a different evaluation from the classifier, visually showing which traits are determinant for the evaluation, and offering us a new method  of augmentation for a fairer model.

The main model is illustrated in Section \ref{Model} while in Section \ref{VAEX} the technique to produce counterfactuals is discussed together with ad hoc improvements of the network.
The counterfactuals samples, obtained from the CelebA Dataset are displayed in Section \ref{Results}, together with quantitative results, whose significance is summarized in Section \ref{Conclusions}.

\section{Model and Architecture}
\label{Model}
\subsection{VAE and reconstruction loss}

Variational Auto Encoders are typically composed of an encoder $\widetilde{q}_\phi(x)$ and a decoder $\widetilde{p}_{\theta}(z)$, which are neural networks trained through self-learning, and they model the sample distribution $P(x)$ by integrating $p(x|z)=p(x|\widetilde{p}_{\theta}(z))$ (the generative density) over a smaller latent space with known density $P(z)$.

The posterior distribution $P_{\theta}(z|x)$ is approximated with $Q_\phi(z|x)$, whose density $q(z|x)$ is inferred by $\widetilde{q}_\phi(x)$, by penalizing the Kullback-Leibler Distance $\mathsf{D}_{\mathbb{KL}}(Q_\phi(z|x)\parallel P_{\theta}(z|x))$. The ELBO results from subtracting the latter from the likelihood objective $\log (p_{\theta}(x))$ and is then maximised through amortized variational inference:
\begin{align*} 
\text{ELBO}: =&\log (p_{\theta}(x))-\mathsf{D}_{\mathbb{KL}}(Q_\phi(z|x)\parallel P(z|x))\\=
\mathbb{E}_{z \sim Q_{\phi}(z|x)}&\big[\log(p_{\theta}(x|z))\big] -
\mathsf{D}_{\mathbb{KL}}(Q_{\phi}(z|x)\parallel P(z)).
\end{align*} 
With independence and Gaussian assumptions, the Log-Likelihood is:
$$\log(p_{\theta}(x|z)) = \sum_i\big[\dfrac{(x_i-\widetilde{p}_{\theta}(z)_i)^2}{2\bar{\sigma}_i^2}+\log(\bar{\sigma}_i)\big]+ C.$$

Following recent ideas~\cite{Asperti}, $\bar{\sigma}^2_i$ is a per pixel variance calculated from the previous batch Reconstruction Error, to which we add momentum for further stability.

\subsection{Hierarchical Models and inference loss}
The independence and Gaussian assumptions that are traditionally attributed to the prior $P(z)$ and the inferred posterior $Q_\phi(z|x)$ are too stringent for expressing complex features and struggle to encode details, hindering sharpness.
Therefore hierarchical models such as~\cite{Sonderby} split the latent variables $z$ in a sequence of layers $(z_k)_{k \leq K}$, whose density endows a normal distribution for $k=0$ or a Gaussian with learned mean and diagonal variance $(\mu_k,\sigma^2_k)= \widetilde{p}_{\theta_{k}^1}(z_{k-1},d_{k-1})$ otherwise, where $d_{k}=d_{\theta_{k}^2}(z_{k-1},d_{k-1})$ is a deterministic path later introduced in~\cite{Maloe}, $d_0$ is a learned parameter~\cite{Vahdat} and  $d_K=\tilde{p_\theta}(z)$.

Similarly, $Q_\phi(z_k|x)$ is a Gaussian inferred by $(\hat{\mu}_k,\hat{\sigma}^2_k)=\widetilde{q}_{\phi_{k}^1}(z_{k-1},d_{k-1},h_{{K-k}})$ (top-down approach), where $h_{j}= h_{\phi_{K-k}^2}(h_{k-1})$ captures high to low level information and $h_{-1}=x$. 
In this setup,
\begin{align*} 
&\mathsf{D}_{\mathbb{KL}}(Q_{\phi}(z|x)\parallel P_\theta(z))=\mathsf{D}_{\mathbb{KL}}(Q_{\phi}(z_0|x)\parallel P(z_0))\\&+ \sum_{1\leq k \leq K}\mathsf{D}_{\mathbb{KL}}(Q_{\phi}(z_k|x,{z_j}_{\{j<k\}})\parallel P_\theta(z_k|{z_j}_{\{j<k\}}))
\\&=
\sum_{k \leq K}-\dfrac{1}{2}-\log(\dfrac{\sigma_k}{\hat{\sigma}_k})+\dfrac{\hat{\sigma}^2_k+(\hat{\mu}_k-\mu_k))^2}{2\sigma^2_k}.
\end{align*} 

This loss encourages the posterior to stay stochastic, and the prior to learn to anticipate the localization of the posterior, given coarsed versions. By doing this the generative model learns hidden features at different stages.

To improve convergence and stability of the loss, in NVAE~\cite{Vahdat} the posterior $\widetilde{q}_{\phi_{k}^1}(z_{k-1},d_{k-1},h_{{K-k}})=\Delta_{\phi_{k}^1}(h_{{K-k}}) +\widetilde{p}_{\theta_{k}^1}(z_{k-1},d_{k-1})$, where $\Delta_{\phi_{k}^1}(h_{{K-k}})$ encodes the information and is trained to  be as small as possible.

\subsection{Architecture}
The architecture is mainly build on blocks [Batch normalization (increased momentum) $\rightarrow$ (hard) Swish activation~\cite{Ramachandran} $\rightarrow$ Convolution or Deconvolution $\rightarrow$ Squeeze and Excitation~\cite{Hu}] influenced by~\cite{Vahdat}, which are added to the residuals. 
Blocks outside the latent space (Fig.~\ref{fig:VAEX_architecture}) have depth of 2. Batch size is $N = 32$ and the Adam optimizer with decaying learning rate is chosen. We prioritize practicality over performance by vastly scaling down the size compared to~\cite{Vahdat}.

\section{VAEX}
\label{VAEX}
\subsection{Enforcing the dependency}
To make  sure that low level information is gathered  by earlier latent variables  we  interpolate  the  residuals  when  changing resolution in $h_k$ and $d_k$ and concatenate the latter with the features prior encoding in  $\Delta_{\phi_{k}^1}(h_{{K-k}},d_{k-1})$.
In the generative path, instead of the commonly used concatenation which can be ignored by the network, we force the dependency between latent variables by inserting an AdaIN~\cite{Huang} layer in $\widetilde{p}_{\theta_{k}^1}(z_{k-1},d_{k-1})$, so that $z_{k-1}$ alters the statistics of the features derived from $d_{k-1}$ immediately before the generation of $z_{k}$. 
This has proven to be an effective way to impact global details (the style)~\cite{Karras} in purely generative models, and combined with the interpolated spatial information, encourages the network to rely on the top latent variable.
We make this even more effective by limiting the pressure in the latent space using the free bits method~\cite{Chen}, of which we introduce a smoothened version: 
$$\bar{\mathsf{D}}_{\mathbb{KL}}=\log(1+e^{\mathsf{D}_{\mathbb{KL}}-\text{FB}})+\text{FB} \quad  \text{with FB}=2,$$ which prevents the posterior collapse and improves tangibly the convergence compared to~\cite{Chen}. 

\subsection{Conditional model}
In order to learn the features associated to the $C$ different classes, we consider the soft-maxed output $\xi=(\xi_c)_{c < C}$ of a classifier of input $x$ as probabilities and we condition $\widetilde{p}_{\theta}(z|\xi)$ and $\widetilde{q}_{\theta}(x|\xi)$ by adding for each $c<C-1$ a constant channel of value $\xi_c$. Similarly, we translate the prior mean of the top latent vector of the $c^{th}$ channel of $s\cdot \xi_c$, where $s=5$ is a scale factor.
By doing so we teach the network to disentangle the desired features associated to each separate class.
The probabilities can be stored in advance, and therefore the model does not need to query the classifier, facilitating dramatically the working pipeline for complex classifiers.
Using probabilities comes with three advantages. Probabilities of an accurate classifier are generally more informative than labels, but they also manifest the bias specific to the classifier. Critically, they can also be used in real life scenarios when true labels are not available.

\begin{figure}[t]
\includegraphics[width=0.5\textwidth]{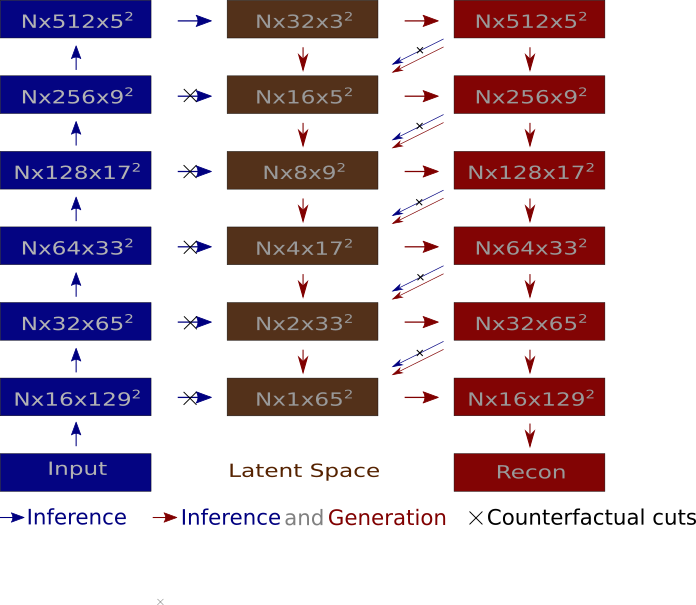}
\caption{Dependencies between features and latent variables represented by blocks together with the features' size. With the exception of the initial one at low resolution, the inference-only connections are cut when $r=0$.}
\label{fig:VAEX_architecture}
\end{figure}

\subsection{Visual Counterfactuals}
\label{Counterfactuals}

The naive approach is to change probabilities associated to the model in favour of the target $c'$ class: $$\widetilde{q}_{\theta}(x|do(\xi_c=\delta_{c,c'}))$$ where we use the $do$ operator \cite{Pearl} and Kronecker $\delta_{c,c'}$. Unfortunately, this approach alone is not often successful, and has little impact on the reconstructed image. Intervening on the $c^{th}$ channel $c<C-1$ of $z_0$ immediately after it is inferred and setting it to $s\cdot \delta_{c,c'}$ also has a limited effect. Nevertheless the same solutions are very effective for the generative model.
\begin{figure}[h]
\includegraphics[width=0.5\textwidth]{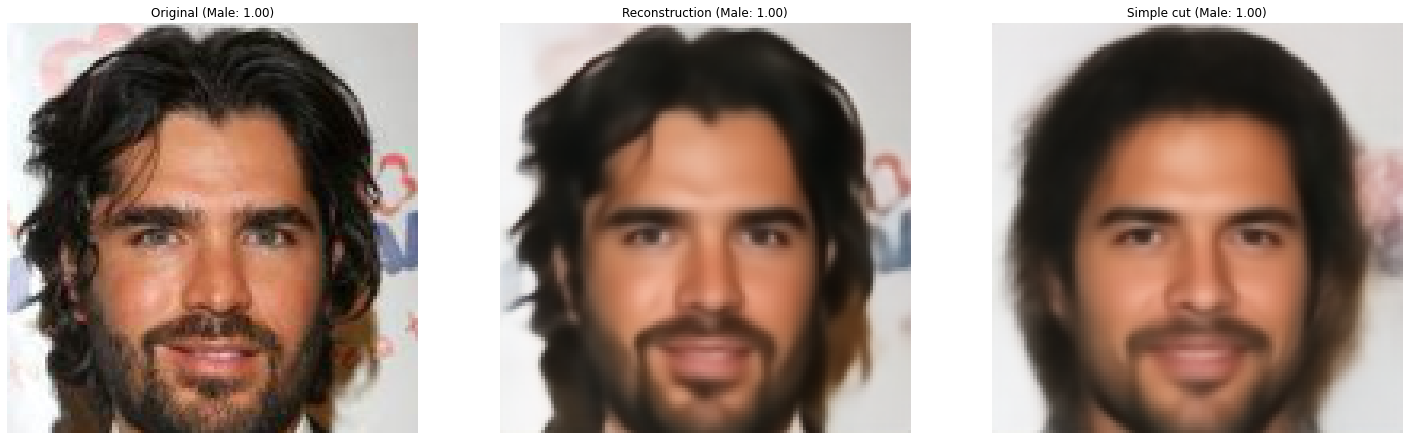}
\caption{Bringing $r$ completely to zero causes a loss of information, which our model seeks to reduce to the minumum.}
\label{fig:simple_cut}
\end{figure}
This is because the posterior of shallow layers is trained to reconstruct the image by pixel loss and pushes the latent variables in tiny regions, with an almost deterministic output. The generative model on the contrary, tries to anticipate the details of the image from the coarser information by learning semantic features, which is what we need to make a meaningful change.
Motivated by this observation, we insert during inference a parameter $r$ such that:
  $$\widetilde{q}_{\phi_{k}^1}=\Delta_{\phi_{k}^1} + r^k \widetilde{p}_{\theta_{k}^1}$$
where $r$ is set to 1 during training. To produce counterfactuals, we condition the model as mentioned and we gradually bring $r$ towards zero, cutting some dependencies (Fig.~\ref{fig:VAEX_architecture}), and relaxing the effect of the posterior on the reconstructed image. When $r=0$ the reconstruction only uses $z_0$, where we have encouraged our network to encode most of the semantic information (Fig.~\ref{fig:simple_cut}). To improve consistency we divide the standard deviation of the latent distributions by $3$. In the Experimental Section we show that usually a partial relaxation is sufficient.

\begin{figure*}[h]
\begin{centering}
\includegraphics[width=\textwidth]{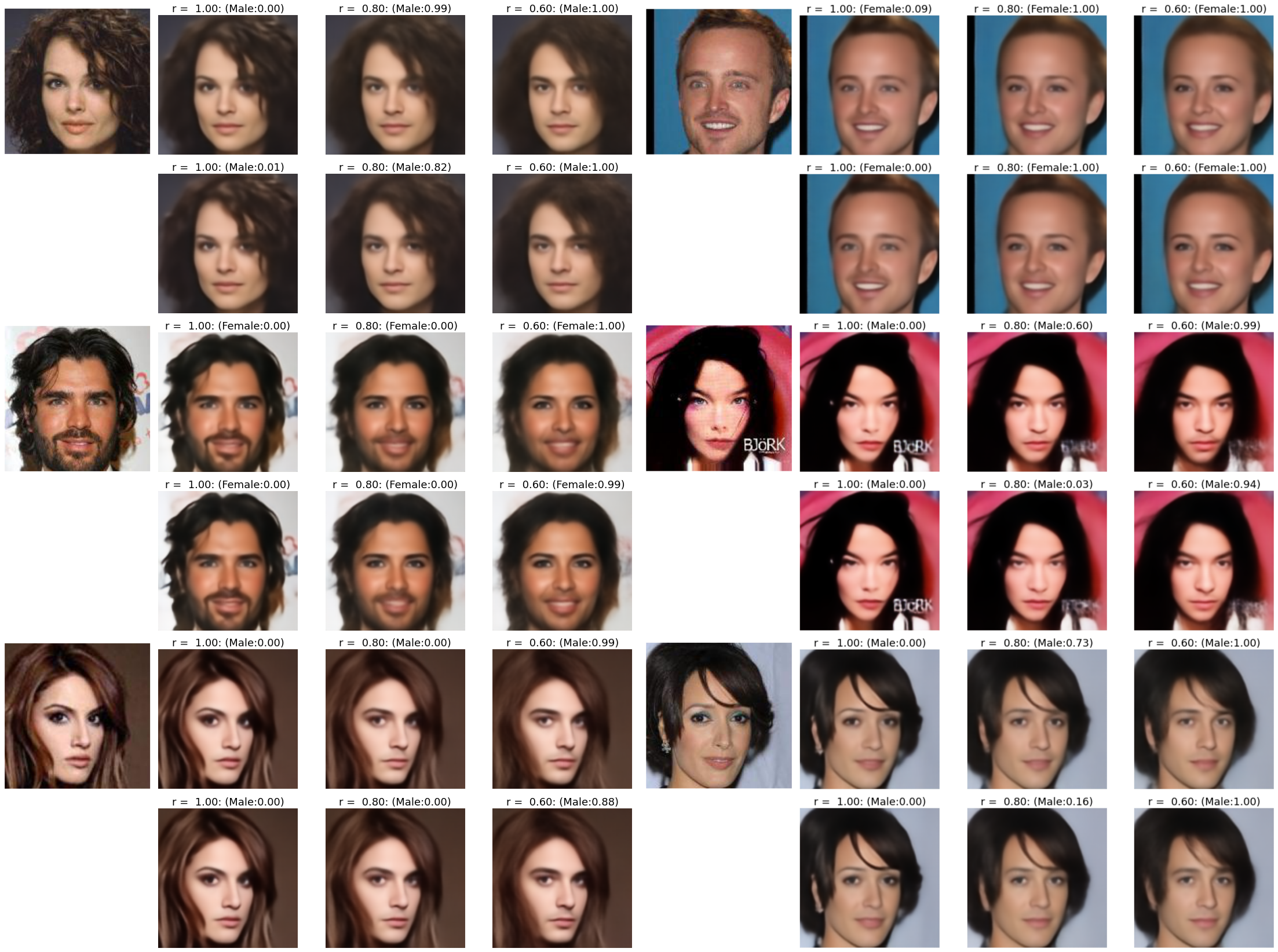}

\caption{ Non-cherry-picked images followed by counterfactuals at different values of $r$. In odd rows VAEX is used while in even rows VAEXcat is used instead. Probabilities are calculated with the investigated classfier. Due to the generative nature of the method, counterfactuals are partially sampled and might present small variations from the ones shown.}

\label{fig:counterfactuals}

\end{centering}
\end{figure*}

\section{Experimental Results}
\label{Results}
In this section we show the efficacy of our method and we compare our model to a version VAEXcat where we use concatenation preceeded by a bottleneck residual block~\cite{Sandler}, similarly to~\cite{Vahdat}. We show the performance in Table~\ref{table:performance}.

Experiments were performed on the CelebA dataset~\cite{Liu} using sex as target label to better compare with~\cite{Barredo}. The dataset is cropped, resized to a square of side 129 pixels and normalized to $[0,1]$.
A simple classifier is trained until it reaches $98.5\%$ accuracy. As the probabilities tended at the extremes, we centered them twice using $f(x) = \dfrac{1}{2}(\sqrt{x}-\sqrt{1-x}+1)$. This allowed the network to train over the $[0,1]$ segment and had a positive effect on the success of counterfactuals.

\begin{table}[t]
\centering

\caption{Test Negative Log Likelihood, Kullback–Leibler divergence, Mean square Error and bits per dimension. }
\label{table:performance}
\begin{tabular}{|c|c|c|c|c|}
\hline
 & \textbf{NLL}&  \textbf{$\mathsf{D}_{\mathbb{KL}}$}& \textbf{MSE} & \textbf{bits/dim} \\ \hline
VAEX  & 103610& 732.5 & 0.0010 & 5.03 \\ \hline
VAEXcat & -104155 & 746.9 & 0.0010 & 5.01\\ \hline
\end{tabular}
\end{table}

The $\mathsf{D}_{\mathbb{KL}}$ is artificially reduced by our method and cannot be used to prove the plausibility of the counterfactuals, therefore we use the FID score \cite{Seitzer}, typically used for GANS, between $2048$ counterfactuals and reconstructions from the test dataset and $2048$ other samples of the test dataset (Table~\ref{table:FID}). As we might expect there is a partial drop in the score when cutting all the connections. In this case we see that the two architectures are comparable.

\begin{table}[t]
\centering
\caption{FID score (counterfactuals produced with $r=0$) }
\label{table:FID}
\begin{tabular}{|c|c|c|}
\hline
 & \textbf{Reconstructions}&  \textbf{Counterfactuals} \\ \hline
VAEX & 48.1 & 66.0\\ \hline

VAEXcat & 49.6 & 66.4  \\ \hline

\end{tabular}
\end{table}

The counterfactuals fool reliably the investigated classifier as shown in Table~\ref{table:Success}. We see that even a partial relaxation is effective most of the time.

\begin{table}[h]
\caption{Percentage of successful counterfactuals varying $r$ }
\label{table:Success}
\begin{tabular}{|c|c|c|c|c|c|c|}
\hline
\textbf{r} & 0 &  0.2 & 0.4 &  0.6 &  0.8  & 1 \\ \hline
VAEX&100.0\%  & 99.8\%  & 99.4\% & 96.1\%  & 71.4\%  & 13.9\% 
\\ \hline
VAEXcat &99.6\% & 99.0\% & 96.9\% & 87.7\% & 54.4\% & 9.8\% 
\\ \hline

\end{tabular}
\end{table}

\label{tab1}

The results above are visually evident from the Fig.~\ref{fig:counterfactuals}: our method when $r=0$, corresponding to a naive conditioning of the posterior, leads to little to none effect while some relaxation of the posterior allows for a full expression of the learned features. We observe that the shift is much more gradual, intuitive and free from unexpected modifications, such as the color of the hair that has been an issue in previous work \cite{Barredo}. At the same time, some counterfactuals manifest some bias of the classifier, which for example expects women to be more smiling. While we can see also here that the classifier is more convinced by the counterfactuals obtained using VAEX than from the ones obtained using VAEXcat given the same support from the posterior, we note that the visual difference is minimal even when the probabilities largely differ (see the last example to the right with $r = 0.8$). We conclude that VAEX is more sensitive to the same features of the investigated classifier. 

\section{ Conclusions}
\label{Conclusions}
In this paper we present VAEX, an hierarchical VAE conditioned to the probabilities outputted by an investigated classifier. Using several architectural solutions we stress the importance of early latent variables, and develop a method of producing realistic counterfactuals, which conserve the resemblance to the original sample but also freely express a semantic alteration without the hindrance of a pixel loss. The model is quick and easy to train, does not require any per sample optimization and does not rely on any label, which makes it attractive to audit a classifier on real life scenarios.

\end{document}